# Advancements in Natural Language Processing for Automatic Text Summarization

Nevidu Jayatilleke
*School of Computing*
*Informatics Institute of Technology*
Colombo, Sri Lanka
nevidu.20200878@iit.ac.lk

Ruvan Weerasinghe
*School of Computing*
*Informatics Institute of Technology*
Colombo, Sri Lanka
ruvan.w@iit.ac.lk

Nipuna Senanayake
*School of Computing*
*Informatics Institute of Technology*
Colombo, Sri Lanka
nipuna.s@iit.ac.lk

*Abstract*— The substantial growth of textual content in diverse domains and platforms has led to a considerable need for Automatic Text Summarization (ATS) techniques that aid in the process of text analysis. The effectiveness of text summarization models has been significantly enhanced in a variety of technical domains because of advancements in Natural Language Processing (NLP) and Deep Learning (DL). Despite this, the process of summarizing textual information continues to be significantly constrained by the intricate writing styles of a variety of texts, which involve a range of technical complexities. Text summarization techniques can be broadly categorized into two main types: abstractive summarization and extractive summarization. Extractive summarization involves directly extracting sentences, phrases, or segments of text from the content without making any changes. On the other hand, abstractive summarization is achieved by reconstructing the sentences, phrases, or segments from the original text using linguistic analysis. Through this study, a linguistically diverse categorizations of text summarization approaches have been addressed in a constructive manner. In this paper, the author explored existing hybrid techniques that have employed both extractive and abstractive methodologies. In addition, the pros and cons of various approaches discussed in the literature are also investigated. Furthermore, a comparative analysis conducted to evaluate the generated summaries utilizing language generation models, employing different techniques and matrices. This survey endeavors to provide a comprehensive overview of ATS by presenting the progression of language processing regarding this task through a breakdown of diverse systems and architectures accompanied by technical and mathematical explanations of their operations. Finally, following the conclusion, a discussion is held regarding the limitations, challenges, and potential future work that can be conducted within this field.

*Keywords— Automatic Text Summarization; Natural Language Processing; Deep Learning; Large Language Models*

## I. Introduction

Automated Text Summarization (ATS) is in high demand due to the exponential growth of textual content across multiple platforms, especially electronic text information. The amount of written material on the Internet and other sources, such as news articles, books, legal documents, medical documents, and scientific research papers is increasing rapidly on an ongoing basis. Therefore, this abundance of information necessitates the need for dependable and efficient advanced text summarizers. Automatic summarization is currently acknowledged as an extremely important task in natural language processing [1].

The objective of automatic text summarization is to extract the key points of the original text without the necessity of reading the complete document. A summary is a concise written composition derived from one or multiple texts, encompassing a significant amount of the information present in the original text [2]. The most significant advantage of text summarization is that it can significantly decrease the amount of time a user spends reading. An exceptional text summarization system should accurately capture the various themes of the document while minimizing repetition [3].

Generating automatic summaries is a challenging endeavor. Summarizing documents requires careful attention to factors such as redundancy, co-reference, sentence ordering, and more, which adds complexity to the task. Since the inception of text summarization in the 1950s, researchers have been striving to enhance techniques for producing summaries that are identical to those created by humans [4].

This study provides a comprehensive analysis of the text summarization task within the field of language processing. It specifically examines various types, classifications, and techniques of summarization in a broader context. Furthermore, this paper extensively examined various methods, techniques, and metrics used for summarization evaluation. This work constructively examines various systems, architectures, and approaches, emphasizing their respective strengths and limitations. Ultimately, the authors conclude the review by acknowledging potential future improvements.

## II. Various Classifications of Text Summarization

Text summarization can be categorized into distinct groups based on various characteristics. The primary factors to consider are the summarization methodology, the number of documents, the algorithm implemented, the domain of the summary, the resulting output summary, the summary style, the summary type, and the language employed.

Abstractive and extractive summarization are the two fundamental approaches that comprise the text summary problem. The objective of extractive summary is to directly extract sentences, phrases, or segments of text from the document without making any alterations. On the other hand, abstractive summaries are produced by reconfiguring the salient sentences, phrases, or sections from the original text through linguistic examinations [5].

Based on the number of documents, ths task can be further classified into two major categories as single-document summarization and multi-document summarizing. Single document summarization involves the construction of a summary based on a single document, whereas multi-document summarization involves the utilization of several documents to construct a sole summary. The activity of condensing an individual document is extended to produce summaries of multiple documents [4].

Considering the style of the summary, text summarization can be categorized again into two types as indicative and informative. Indicative summaries offer a summary of the content included within a given document. The information they offer is applicable to the topic addressed in the document. Informative summaries offer extensive and complex information regarding the subjects they encompass [4].

Language-wise, summaries can be classified into three distinct categories as cross-lingual, monolingual, and multilingual. A monolingual summarization system is characterized by the linguistic identicality between the source document and the target document. A multi-lingual summarization system is characterized by the provision of source documents in various languages alongside the generation of summaries in corresponding languages. A cross-lingual summarization system is a system that produces a summary in a chosen language other than the source document's lingo [4].

Summaries may unveil either a generic or query-centric makeup, accustomed upon their components. Query-focused summaries, alternatively referred to as topic-focused or user-focused summaries, are summaries created to specifically address the user's query or topic of interest. Query-focused summaries are designed to specifically emphasize the content that is important to the query at hand, whereas a generic summary offers a broader overview of the information contained within the document [4][6].

Summaries can be grouped into two more distinct groups based on the specific domain they relate to: general or domain specific. The general, or domain agnostic ATS system is responsible for summarizing documents from diverse fields. On the other hand, the domain specific ATS system is specifically engineered to consolidate papers within a distinct field, such as medical, legal, or academic records [4].

Furthermore, it is possible to classify summaries into four distinct categories based on their type as headline, sentence-level, highlights, or full summary. In accordance with the intended function of the ATS system, the size of the manufactured summaries varies. The process of generating headlines generally results in crisp headlines that are shorter than a complete sentence. The process of sentence-level summarization involves generating a concise and generally abstract sentence based on the provided input text. Typically presented in telegraphic bullet points, a highlights summarization produces an informative and condensed summary. The summary of highlights delivers the consumer with an understandable summary of the essential insights presented in the input document(s). In general, the generation of a full summary is determined by either the desired length of the summary or a reduction factor [4].

Importantly, text summarization techniques can be broadly classified as unsupervised learning and supervised learning, based on algorithms and available data. For the implementation of supervised algorithms, it is necessary to have annotated training data available during the training period. The process of constructing training data necessitates the manual annotation of data by human operators, thereby presenting a formidable and expensive undertaking. The unsupervised algorithms, in contrast, does not require the use of annotated training data [3][4].

It is possible to breakdown text summarization task further into different categories. However, these mentioned types are the most pertinent of them all and many of the conducted research on text summarization lies within these classifications. In this study, the author will focus on the important fundamental classification of text summarization: abstractive and extractive based on the broader context of other classifications.

### III. Extractive Text Summarization Approaches

The discipline of Natural Language Processing (NLP) inaugurated the development of extractive summarizing significantly earlier than abstractive text summarization. There exist both supervised and unsupervised learning methodologies that demonstrate efficacy in addressing this task. Significantly, statistical methodologies, including machine learning and deep learning approaches, have demonstrated their utility in extractive text summarization. This is elaborated upon in the following subsections.

*A. Graph Based Methodologies*

Graph-based methods employ graph theories to construct techniques for summarizing text. The sentences within the documents are depicted as nodes in a directed or an undirected network using typical preprocessing techniques such as stemming and stop word removal. Sentences are connected by edges that are determined by the structure of the sentence [6]. In LexRank and TextRank, this type of representation is commonly employed for the purpose of extractive summarization [7].

The construction of a graph is necessary for TextRank, wherein the vertices correspond to the sentences that are to be ranked. The determination of sentence "similarity" is based upon the degree of overlap in their respective content. The graph exhibits a high degree of interconnectivity, wherein each edge is imposed a weight that signifies the magnitude of ties between distinct pairs of sentences within the given text. The text is depicted in the form of a weighted graph, and subsequently, a ranking process based on weighted graphs is conducted. After the graph is processed by the ranking algorithm, sentences are organized in a decreasing order according to their score. The sentences with the top scores are selected for insertion in the summary [8].

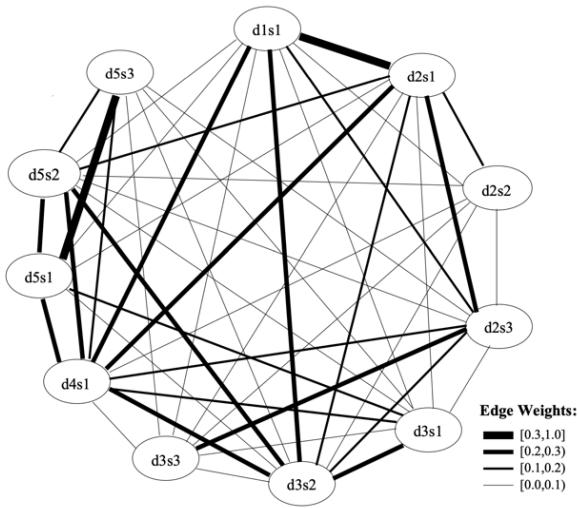

Fig. 1. Weighted cosine similarity graph [9].

The LexRank algorithm is likely to provide importance to a sentence that showcases similarity to many other sentences within the given text. The approach entails suggesting a certain sentence by assessing its resemblance to other sentences, leading to an elevated ranking. The methodology employed in this study is founded upon the concept of Eigenvector Centrality and follows an interconnected graph structure. As illustrated in Fig. 2, every sentence is situated at a vertex of the graph, and the weights assigned to the edges are calculated utilizing a cosine similarity metric [9].

The LexRank algorithm draws inspiration from the TextRank algorithm and can be considered as an enhanced version of the TextRank algorithm. There exist notable distinctions between the two algorithms, which are subsequently accompanied by shared characteristics. Nevertheless, there are notable distinctions in the ranking of sentences, particularly in the utilization of cosine similarity to assess word overlaps across phrases in LexRank compared to TextRank. Both methods operate on undirected graphs, which is a significant commonality.

These algorithms continue to be employed and deliberated in scholarly articles, in addition to their application in hybrid methodologies which consists of their own pros and cons.

Most importantly the graph algorithms are domain independent and language independent [7]. The extraction of summaries requires minimal resource consumption. Also, it improves the coherency and identifies unnecessary repetition [3].

However, they operate on the assumption that the weights of all words same, hence disregarding the significance of individual words within the document. Also, graphs that represent sentences as Bag of Words and using similarity metrics may not be able to recognize semantically identical sentences [7]. Furthermore, it does not address concerns such as the problem of dangling anaphora [3].

### B. Machine Learning Algorithms

Researchers strongly prefer machine learning rooted methods for extractive text summarization among the current approaches. The ML algorithms' prominence in text-generating tasks emanates from their scalability, enabling them to effectively process substantial volumes of data. The availability of supervised and unsupervised learning algorithms in ML techniques makes it very adaptive to the evolving nature of challenges.

Supervised learning techniques offer the ability to transform the challenge of text condensation into a supervised classification process at the sentence tier. The system uses a training set of documents to acquire knowledge on how to classify every sentence in the evaluation document as either 'summary' or 'non-summary' by utilizing examples. Sentences categorized as summary can be combined to create the final summary [10]. This pertains to a binary classification problem, which involves the classification of data into two separate classes [7].

The predominant methodology employed in unsupervised learning entails the implementation of clustering algorithms. Various algorithms for clustering are implemented for the purpose of sentence extraction.

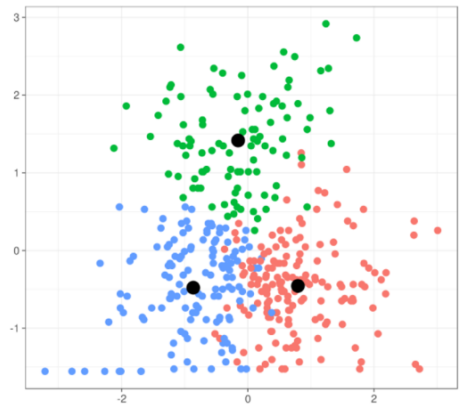

Fig. 2. Clustering semantically identical sentences.

K-means is an unsupervised machine learning clustering algorithm that divides a large input data set into k distinct groups, with each group being represented by the average of data points inside that group, known as centroids, as shown in Fig. 2. The centroids symbolize the arithmetic mean of their clusters, which are the important concepts inside their collection of sentences. The method utilizes these centroids to choose sentences, as they are calculated as the arithmetic mean of the clusters of sentences, rather than a specific sentence. The objective is to choose sentences that have the shortest Euclidean distance to the centroids [11].

Machine learning algorithms are widely recognized as effective methods in the domain of extractive text summarization, demonstrating acceptable results that can compete the more complex architectures discussed in existing literature. Furthermore, the incorporation of these algorithms into composite models for generating abstractive text summaries in a hybrid context is a pertinent subject that demands deeper investigation.

Importantly, machine learning algorithms are adaptable to different languages and domains with relative ease. Their capability to interpret the procedural steps followed to generate a decision, which results in conveying them as white box models make them much comprehensible.

However, they require a dataset of manually manufactured extractive summaries such that every sentence in the novel training documents can be designated as either ''summary'' or ''non-summary'' when leveraging classification [3]. These algorithms also may struggle to capture and model linguistic variabilities effectively.

*C. Deep Learning Methodologies*

Deep learning methodologies, which are extensions of machine learning algorithms, exhibit increased architectural complexity due to the incorporation of neural networks and their variants, which produce superior results.

A novel approach is proposed for creating concise and reliable summaries of lengthy documents by combining **Restricted Boltzmann Machine** (RBM) and **fuzzy logic**, together with set operations. The Restricted Boltzmann Machine (RBM) is a neural network model that comprises of two layers: an input layer and a hidden layer. By utilizing RBM, it becomes apparent that it appoints a score to every sentence in a document, and these scores are used to generate a summary (summary #1).

Next, the sentence scores are fuzzified using triangle membership functions within the framework of fuzzy logic. The fuzzification categorizes the scores into three levels: High, Medium, and Low. De-fuzzification is then employed to ascertain the significance of the statement, considering a set of well-defined criteria. The technique identifies essential sentences and combines them to create another summary (summary #2). The procedure for obtaining the final summary involves identifying both common and uncommon set of sentences from the first and second summaries. The Final summary includes the common set of sentences only [12].

A different approach entails constructing a matrix that represents the features of sentences, and then utilizing AutoEncoder Networks to enhance the matrix to precisely score the sentences. To train the summarizer in the current study, features were included by extracting them from the raw data. These traits allowed the researchers to acquire sentences that included more significant information in the text. These features combine to create a vector referred to as a feature vector. Every sentence in the text was associated with a feature vector.

Autoencoders have output layers that are similar to the input layers, and their objective is to accurately reproduce the input layer values in the output layer. The researchers employed a neural network consisting of nine layers, with seven of them being hidden layers. This network consists of two components: an encoder and a decoder. During the encoding stage, the network efforts to transform the data and create new features based on the features it has received as input. Within a decoder network, the network regenerates the input data using the features generated in the preceding stage. The network outputs include of sentences with updated scores. The sentences were arranged according to their scores and then picked based on the applied word restriction for the summary [13].

Deep learning models provide significant computational capabilities, but they also exhibit advantages and disadvantages that can be evaluated. These models are highly scalable, enabling them to effectively process substantial volumes of data. Also, they are capable of handling complex linguistic structures due to more detailed feature representations [14].

However, they may struggle to capture and model linguistic variabilities effectively and requires large datasets to train models to learn feature representations and due to complexities of the architectures. Furthermore, it is difficult to define the procedure followed to generate a decision which depicts them as black box models.

Prior to the emergence of transformer architecture, the field of text summarization was mainly concentrated on extracting sentences or phrases rather than using generative AI for abstractive text summarizing [15]. This study provided a glimpse into the evolution of extractive text summarization across different methodologies among the plethora of available research.

## IV. ABSTRACTIVE TEXT SUMMARIZATION APPROACHES

Abstractive text summarization, as in contrast to extractive text summarization, generates new text from input, which signifies its entry into Generative AI. As stated in section II, abstractive text summarization involves constructing a summary that resembles that of a human while considering the significance of sentences and phrases within the input text. This method usually involves making significant modifications and rephrases to the original sentences to achieve the intended summary [16].

*A. Conventional Deep Learning Methodologies*

The current body of research on abstractive text summarization largely use neural network architectures, as they possess a robust capacity to comprehend textual context and produce original text content.

The **Seq2seq** framework, often referred to as the encoder-decoder framework, is widely acknowledged as the most effective methodology for translating text from one format to another, such as speech recognition, question answering systems, and machine translation, etc. These models utilize an encoder to recognize, comprehend, and analyze the input sequence, and utilize the multi-dimensional dense feature vector to describe it. Subsequently, the feature vectors of the input items are utilized on the decoder side to systematically produce the output items. Fig. 4 illustrates the fundamental structure of the encoder-decoder system [17].

The encoder-decoder framework is the fundamental and central foundation of deep learning rooted abstractive text summarization models. The encoder and decoder are developed utilizing diverse neural networks. This section focuses on conventional neural networks that utilize well-established and frequently used deep learning architectures within an encoder-decoder framework.

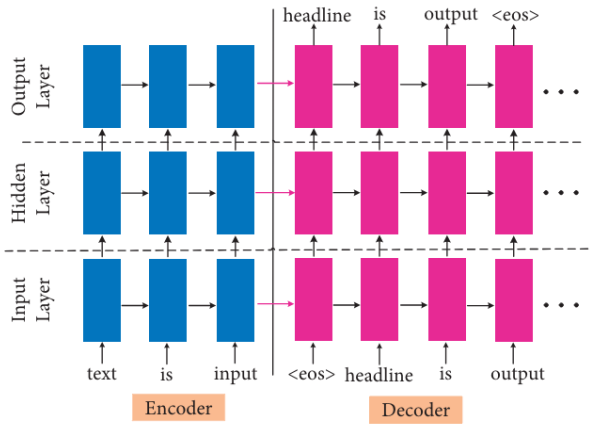

Fig. 3. The basic encoder-decoder framework [17].

A novel approach is described that utilizes a standard Feed-Forward Neural Network (FFNN) language model with an attention mechanism within an encoder-decoder framework. The primary attention-based encoder was introduced alongside convolutional and bag of words encoders for the purpose of research benchmarking. The text generation component included a beam search decoder that maintains the entire vocabulary V while limiting itself to K possible hypotheses at every position of the summary. This marks the inaugural integration of deep learning into abstractive text summarization, signifying a significant milestone in this field [18].

A different method that employs Recurrent Neural Networks (RNN) adopting the Seq2seq framework was proposed for abstractive text summarization. In the RNN encoder-decoder framework, the encoder component combines the vector mapping of the current input word with the output of the hidden states of all prior words at specific hidden states. This combined information is then passed on to the next hidden state. The input string is processed by the encoder, and the resulting output from the last hidden state is used as a vector called the context vector, which is then sent to the decoder. Alongside the context vector, which is inputted into the initial hidden state of the decoder, the start-of-sequence symbol was fed to construct the first word of the summary based on the headline. Every word that is generated is used as an input for the subsequent decoder hidden state to create the subsequent word of the summary. The final word produced is represented by the end-of-sequence sign. Prior to constructing the summary, every output from the decoder will be transformed into a distributed representation. This representation is then sent via the softmax layer and attention mechanism to construct the resulting summary [19][20].

When the sequence becomes excessively long, Recurrent Neural Networks (RNN) start to exhibit the phenomena of gradient explosion and vanishing. In contrast to recurrent neural networks (RNN), Long Short Term Memory (LSTM) [21] effectively addresses the issue of long-term dependencies by selectively retaining information via input, forget, and output gates. A proposed strategy involves utilizing an Encoder-decoder model that incorporates Bi-LSTM (a variant of LSTM) along with an attention mechanism. This approach aims to enhance the accuracy of generating contextual phrases and mitigate issues related to repetition. In this case, the encoder employs a Bidirectional Long Short-Term Memory (Bi-LSTM) model with a self-attention mechanism, while the decoder utilizes a unidirectional LSTM model with beam search. This methodology involves identifying crucial elements, assessing the context, and renewing them. This guarantees that the most crucial information is communicated using the fewest possible words [22].

Furthermore, a suggested framework for abstractive text summarization utilizes a synthesis of LSTM and Convolutional Neural Network (CNN) to generate new sentences by investigating smaller, more detailed units identified as semantic phrases rather than entire sentences. The model architecture consists of a single-layer CNN encoder and an LSTM decoder. The system consists of two primary stages, the initial stage retrieves phrases from the source sentences, and the second stage produces text summaries using deep learning. Additionally, the approach separates the decoder component into two distinct modes: generate mode and copy mode. During the generate mode, the decoder predicts the next phrase, while the copy mode identifies the position of the current phrase in the original text and copies the following phrase into the summary. Thus, this can be characterized as a hybrid strategy because it incorporates elements of both extractive and abstractive methods. Nevertheless, the ultimate summary is primarily a result of abstractive text summarizing [23].

Abstractive text summarization is a task that can be accomplished with Generative Adversarial Networks (GANs) owing to their capability of producing features, learning the entire sample distribution, and generating correlated sample points. The GAN model depicted in Fig. 4 comprises of three components. The first component is a generator that consists of an encoder and a decoder, both of which are two-layer models based on LSTM. The second component is a four-class classifier known as a similarity discriminator. The similarity discriminator consists of two encoders, both of which are CNN-based models. These models follow a specified order of convolution, max pooling, and activation layers. Another CNN-based model in the proposed GAN architecture is the readability discriminator, serving as the third component. It indicates if the summary was created by the generator or a human. The generator is further enhanced through the utilization of a policy gradient technique, so transforming the problem into the domain of reinforcement learning [24].

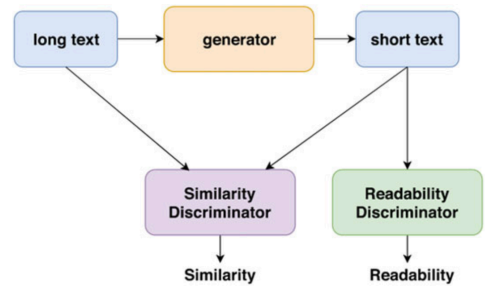

Fig. 4. GAN Architecture for Abstractive Text Summarization [25].

A dual encoding approach utilizing a Seq2seq RNN was proposed, comprising a primary encoder, a secondary encoder, and a decoder endowed with an attention mechanism. All the aforementioned modules utilize the Gated Recurrent Unit (GRU). A GRU based RNN computes semantic vectors for every word in the input sequence as its primary encoder. The secondary encoder initially computes the significance weight for every word in the input sequence and subsequently recomputes the associated semantic vectors. In contrast to the primary encoder, the secondary encoder is constructed using a unidirectional GRU. The secondary encoder is employed in the sequence generating task as a complementary and independent encoder to enhance the effectiveness of our fundamental model. The decoder, which utilizes a GRU with an attention mechanism, decodes the input in a step-by-step manner, generating a partial output sequence of a fixed length at each stage [25].

The attention-based encoder-decoder models in recurrent neural networks have shown impressive results in short-text summarization undertakings. However, these attention encoder-decoder models frequently encounter unforeseen downsides of producing repeated words or phrases and lacking the capacity to handle Out of Vocabulary (OOV) words effectively. To tackle these concerns, a proposed strategy involves incorporating a joint attention mechanism into the output sequence, as depicted in Fig. 5. This mechanism aims to prevent the inclusion of repetitive content and handles uncommon and unfamiliar words through the subword method. The presented study employs the subword model to address the problem of uncommon and OOV words by dividing words into smaller subword units. This approach offers the benefit of simplifying the summarization procedure and minimizing the training required, while still achieving accuracy comparable to models that use a larger vocabulary [26].

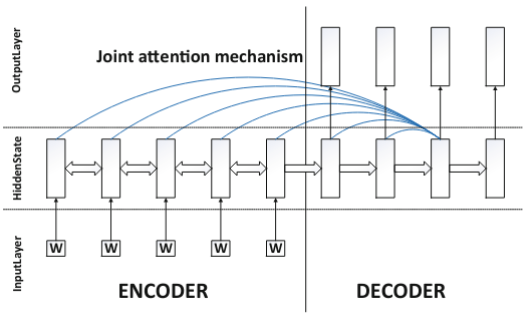

Fig. 5. *Neural model of joint attention.* Joint attention refers to the simultaneous focus on both the input sequence and the output sequence. The attention on the input sequence is utilized to retain and provide more comprehensive details about the input. The attention on the output sequence is utilized to prevent the repetition of phrases [26].

The systems presented primarily used encoder-decoder and GAN frameworks, which were implemented using FFNNs, RNNs, LSTMs, CNNs, and GRUs. These conventional deep learning approaches marked the inception of successful abstractive text summarization. The inclusion of attention mechanisms was an important attribute that greatly contributed to the enhancement of model performances. More importantly, these methods serve as the basis for transformer-based **Large Language Models** (LLMs), which is further explored in the subsequent subsection.

These conventional deep learning models are proven to have enhanced semantic coherence compared to traditional approaches [27]. Also, they encode textual information into latent embedding spaces with high efficiency [28] which provides more clarity during model training.

However, there are challenges in copying and recalling information despite the use of LSTMs and GRUs [29]. Most commonly, they are resource intensive and comprises training inefficiencies [30].

*B. Graph Based Methodologies*

Graphs are frequently employed for extractive summarization, where the graph is typically undirected and consists of texts as nodes connected by edges representing similarity. Here, the graph data structure is a unique technique where each node represents a word unit, and the directed edges depict the sentence structure. For this task to represent natural language text, the central concept is to represent it using a graph data structure known as Opinosis-Graph and to formulate this abstractive summarization problem as locating ideal paths within the graph. This formulation exhibits characteristics of abstractive summarization by incorporating fusion (the combination of extracted components) and compression, distinguishing it from traditional sentence level extractive summarization.

To accomplish the desired outcome, several sub routes in the graph are examined and evaluated according to the following scoring system: Assign a numerical rank to each of the paths, and then arrange their scores in a decreasing order. The ranking also encompasses the collapsed paths. Remove redundant (or highly similar) pathways by use of a similarity metric such as Jaccard. Choose the most prominent residual paths as the created summary, with the number of paths to be selected determined by a value that specifies the desired summary size [7][31].

Another suggested approach is the utilization of Gated Graph Neural Attention Networks (GGNANs) for abstractive summarization. The renowned Seq2seq and the proposed GGNANs unified graph neural network are utilized to encode graph-structured data more efficiently. The model comprises two components: a sentence encoder and a decoder. The encoding component consists of a word embedding layer, a two-layer Bi-LSTM, and a GGNN. The decoder consists of a unidirectional LSTM and a sequence attention mechanism. Formally, a directed graph is a mapping of a set of nodes and edges. The suggested approach establishes connections between each node based on the correlations of word co-occurrence in the entire text. To incorporate global graph-structured data, they utilize a sliding window of a predetermined size across the text to calculate the information on word cooccurrence. They further employ a technique called point-wise mutual information (PMI) to compute the importance between two words. To preserve the sequence-structured information in the graph, they incorporated self-connections, forward connections, and backward connections.

By constructing a graph, they were able to effectively unite graph-structured and sequence-structured information [32].

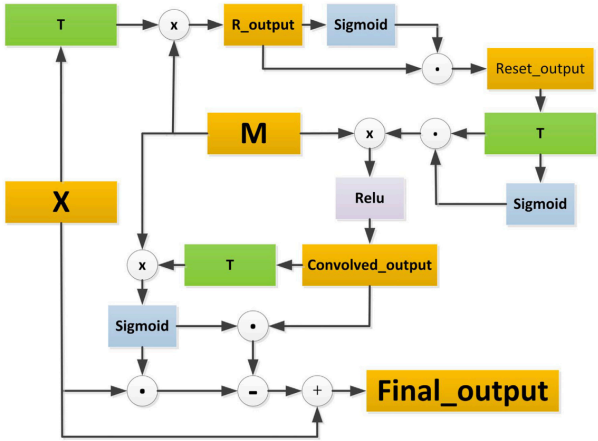

Fig. 6. *The internal architecture of GGNNs*. The X denotes the hidden mapping of the Bi-LSTM, T indicates the process of matrix transformation, and the M depicts the incidence matrix [32].

These graph based models are capable of capturing complicated relationships and dependencies among words, sentences, or documents [33]. Comparably easy to construct and understand the procedure end to end. Importantly, they facilitate the incorporation of diverse linguistic features and semantic information [34].

Unfortunately, their outcome is greatly influenced by the quality of the graph structure and its way textual units are represented within the graph [35]. Also, these structures are computationally complicated when involved in processing massive graphs [36].

Graph based techniques are highly efficient and have made significant advancements in several tasks, including those associated with text, despite being less prominent. Still, the multiple variations of GNNs contribute significantly to the advancement in the field of text generation.

### C. Transformer Based Methodologies

Transformers are the architectures which hold the core elements of LLMs, which have gained significant popularity in recent years for tasks involving text generation. The emergence of this architecture, which was progressively offered to humans via different systems, possibly demonstrated the extensive potential of AI to the highest extent. This study explores the application of several fine-tuning and transfer learning [28] approaches on language models to undertake abstractive text summarization.

Transformer too follows an encoder-decoder structure using stacked self-attention and pointwise, fully connected layers for both the encoder and decoder, as shown in Fig. 7 [15]. The original transformer architecture can be depicted as follows:

- Encoder: The encoder consists of a stack of 6 equivalent layers. Every layer is equipped with a multi-head self-attention mechanism and a basic, position-wise fully connected feed-forward network. A residual connection is used for the two sub-layers, followed by layer normalization [15].
- Decoder: The decoder is also consisting of a stack of 6 similar layers. The decoder incorporates a third sub-layer that, alongside the existing two sub-layers, applies multi-head attention to the output of the encoder stack. Like the encoder, the residual connection is employed around the two sub-layers, and it is then followed by layer normalization. To avoid positions being influenced by following positions, the decoder utilizes a modified self-attention sub-layer.
- Attention: An attention mechanism can be defined as mapping a query and a set of key-value pairs to an output, where the query, the keys, the values, and the output are vectors. The output can be determined by computing a weighted sum of the values. The weight assigned to every value is generated using a compatibility function that compares the query with the corresponding key.

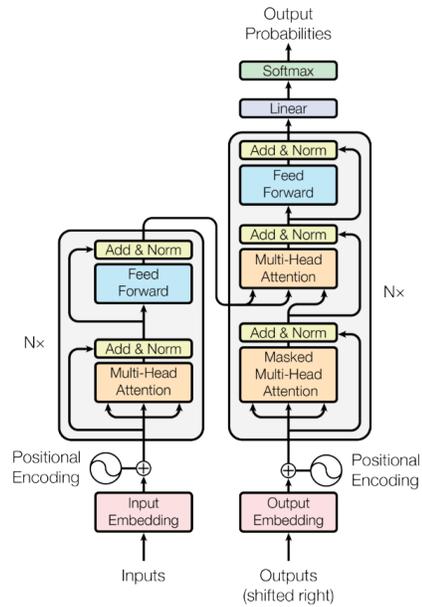

Fig. 7. The Transformer Model Architecture [15]

A proposed approach involved implementing abstractive text summarizing on reviews by fully fine-tuning the **Bidirectional Auto-Regressive Transformer** (BART) architecture. This implementation considerably enhances the model and improves the overall quality of summary. Full fine-tuning of a pretrained model involves changing all the weights of the model parameters during the training phase. To facilitate the enhancement of the input data, the tokenization process employs BartTokenizerFast. In addition, the model employed a sampling technique called Sortish sampling to choose tokens, resulting in improved smoothness and speed. Moreover, the utilization of weight decay boosts the performance of the model through the addition of regularization [37].

A different approach consists of two unique trainable components. An extractive model, comprises a hierarchical encoder that produces sentence representations, used to classify sentences in the input and a transformer model, accustomed on the extracted sentences and a part of or the complete input document. A single transformer language model with 220M parameters, 20 layers, 768-dimensional embeddings, 3072-dimensional position-wise Multi-Layer Perceptron (MLPs), and 12 attention heads has been constructed and trained from scratch. To enable an unconditional language model to perform abstractive summarization, they had used the certainty that language models are trained by decomposing the joint distribution of words in an autoregressive manner. Put simply, they usually factorize the joint distribution of tokens into a product of conditional probabilities. Consequently, they structure the training data for the models in a way that ensures the ground-truth summary follows the knowledge utilized by the model to produce a summary [16].

A novel approach proposed PEGASUS, a Seq2seq model with gap-sentences prediction as a pretraining goal designed for abstractive text summarization. After deliberating on various gap-sentence selection techniques, they concluded that principle sentence selection is the most effective approach. Within the PEGASUS framework, significant sentences are selectively eliminated or masked from an input document, and subsequently created into a single output sequence alongside the residual sentences, like an extractive summary. As a pre-training target for subsequent summarization tasks, they discover that masking entire sentences from a document and producing gap sentences from the remainder of the document is effective. By employing their most optimal 568M parameter model, they achieve results that are equivalent to or surpass the state-of-the-art on 12 downstream datasets that were regarded at the time of publication [38].

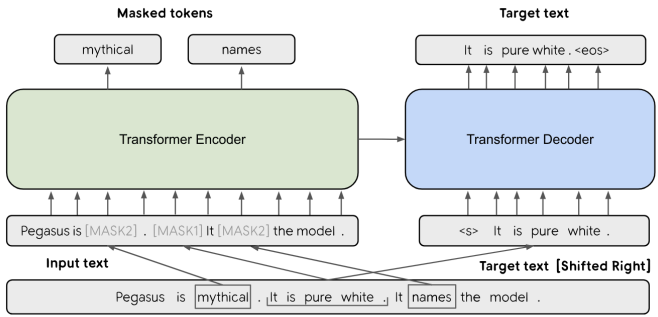

Fig. 8. The fundamental structure of PEGASUS is a conventional Transformer encoder-decoder architecture. Both GSG and MLM are employed concurrently in this example as pre-training goals. Initially, there are three sentences. One sentence is concealed with [MASK1] and employed as the target generation text (GSG). While the input retains the other two sentences, certain tokens are randomly masked using [MASK2] (MLM) [38].

An alternative methodology introduced a novel model known as LongT5, which was utilized to examine the simultaneous effects of modulating the input length and model size. To accomplish this, they include the concepts of long-input transformer attention and pre-training into the scalable T5 model [39] architecture. The main concept is to selectively mask salient sentences from a document and urge the model to generate them as a single string, as if it were a summary. The key distinction between T5 and LongT5 is in their attention mechanism from an architectural perspective. They explore two types of attention mechanisms for LongT5: Local Attention and Transient Global Attention (TGlobal). Both variations maintain multiple characteristics of T5, including relative position representations, support for example packing, and compatibility with T5 checkpoints. The enhanced performance of this pretrained model, which was fine-tuned with multiple datasets, was primarily due to its capacity to process lengthy inputs, which makes it well-suited for summarization tasks [40].

A patent document summarizing framework is proposed, which includes a learning stage based on a GAN architecture. The GAN-based summarization model consists of a generator and a discriminator that vie with each other. The generator is a transformer-based model for text summarization. It produces summary sentences by taking in individual patent textual information and label data as input. The generator comprises of an encoder and a decoder. The construction of the transformer involves setting up of encoder and decoder blocks in a stacked arrangement. For the present study, a total of four blocks were piled and utilized. The generator's encoder block comprises an embedding layer, a multi-head attention layer, a regularization layer, a feed forward layer, and another regularization layer. Upon inputting the complete text into the encoder, the layer proceedings are executed in a sequential manner, resulting in the calculation of the encoder block output. At this point, the transformer model applies positional encoding to merge word embedding and location data of each word. The decoder comprises a masked multi-head attention layer, an initial regularization layer, a multi-head attention layer, a second regularization layer, a feed forward layer, and a final regularization layer. The word produced in the preceding stage is fed into the first layer of the decoder, while the output of the encoder block is fed into the multi-head attention layer of the decoder block. The output of the decoder block is ultimately fed into the dense layer, which then generates the next word.

The discriminator is a model that takes in the textual information of summary sentences and assesses if each sentence is a created or an actual sentence. The discriminator in this work is a classification model that has a single Bi-LSTM layer. It obtains the created summary sentence or the target summary sentence as input and verifies its authenticity, distinguishing between actual and generated sentences. The discriminator's output serves as a reward to train both the generator and discriminator [41].

Another study adopts a fresh approach that involves pre-training a large transformer model and then specializing it for text summarization. The research showcases the model's capability to efficiently generate summaries. To optimize the utilization of pre-trained weights, they employ a network that consists only of a Transformer-based decoder during the process of fine-tuning [42]. The decoder-only network, approaches summarization as a language modeling task in which a summary is appended to each article example. Instead of employing distinct encoder and decoder modules, a unified network is utilized to both encode the input and produce the output. Importantly, it contains pre-trained self-attention parameters that are utilized to focus on both the original and the formerly produced target representations. This method negates the redundancy of appointing duplicates of the same pre-trained weights into both the encoder and decoder. It also utilizes fewer parameters in comparison to encoder-decoder networks. Most crucially, it guarantees that all model weights, including those that govern attention over original states, are pre-trained [43].

Transformers are exceptionally effective at capturing long-range dependencies across various sections of text and their context, which allows them to provide summaries that are more accurate and logically connected to the context [44]. Certain summarization tasks might benefit from the fine-tuning of pre-trained transformer models, which enables effective transfer learning and enhances performance on downstream tasks [45]. Also, transformers are highly scalable and can manage massive datasets efficiently, making them ideal for handling large amounts of text data for summarization [46]. Most importantly, these models have achieved state-of-the-art summarization tasks, showcasing exceptionally good summaries [47, p. 20].

However, transformer models are often criticized for their lack of interpretability, which poses a challenge in comprehending the model's decision-making process [44]. Also, they require large amounts of training data to achieve best performance, which can provide a challenge in situations when there is a scarcity of annotated data [27]. Transformer models commonly are highly resource intensive and demand substantial computing power for both training and inference [48]. Furthermore, adapting transformer models to specific domains or languages may necessitate additional training and fine-tuning efforts, which can be demanding in terms of resources and time [27].

Currently, the transformer-based language models excel in text generation tasks. Factors such as their attention mechanism, bidirectional context modeling, transformer architecture, pre-trained language models, fine-tuning capabilities, multimodal extensions collectively contribute to its success.

## V. Evaluation Methods in Text Summarization

Assessing the excellence of a summary is a highly challenging operation. Significant queries remain on the suitable techniques and categories of assessment. Multiple criteria can be used to evaluate the performance of summarization systems. A system summary can be compared to the source text, a human crafted summary, or another system summary. Summarization evaluation methods can be categorized into two major types. Extrinsic evaluation assesses the quality of summaries by evaluating their usefulness for a specific task, while intrinsic evaluation directly analyzes the summary itself to determine its quality [49].

The intrinsic evaluations can be classified into two primary categories: content and text quality evaluation. This study examines various content-based criteria that are routinely used to assess text summaries. Typically, while evaluating results, the common strategy is to verify if the output matches the actual target precisely. However, in text-generation tasks, it is impractical to adhere to that strategy because the resulting sequence will often not exactly match the reference. Thus, methods to calculate the similarity between the created and actual summaries are employed.

Various distance metrics can be directly utilized for the task of similarity learning. The greater the distance between two patterns, the lower their similarity. The Euclidean distance (1) is widely recognized as the most often employed and straightforward distance measure. In the context of Euclidean space, the distance between two points is defined as the magnitude of the line segment that directly connects them. The Euclidean distance, denoted as d(x, x'), is defined as the distance between two patterns x and x' using the Euclidean metric [50].

$$d(x, x') = \sum_{i=1}^{d}(x_i - x'_i)^2, \quad (1)$$

The probability of similarity reduces as the Euclidean distance increases.

Cosine similarity (2) is a frequently employed metric, especially in high-dimensional positive spaces, for activities like information retrieval and data mining. Compared to the Euclidean distance, which is highly sensitive to even slight deformations, cosine similarity places greater emphasis on directions. The similarity is measured by calculating the cosine of the angle between two vectors. It is anticipated that two vectors that are similar will have a small angle between them. The cosine similarity between two vectors x and x' is defined as follows [39]:

$$\cos(\theta) = \frac{\sum_i x_i x'_i}{\sqrt{\sum_i (x_i)^2} \cdot \sqrt{\sum_i (x'_i)^2}}, \quad (2)$$

where $\theta$ represents the angle between x and x'. The similarity between these patterns increases as $\cos(\theta)$ increases.

The most popular similarity metric used for summarization tasks is Recall-Oriented Understudy for Gisting Evaluation (ROUGE). It involves multiple automatic evaluation methods, such as ROUGE-N (3), ROUGE-L, ROUGE-W, and ROUGE-S, which determine the similarity between summaries. The most prevalent among them is the ROUGE-N detailed below:

$$ROUGE - N = \frac{\sum_{S \in \{RS\}} \sum_{gram_n \in S} Count_{match}(gram_n)}{\sum_{S \in \{RS\}} \sum_{gram_n \in S} Count(gram_n)}, \quad (3)$$

Where n denotes the length of the n-gram, $gram_n$, and $Count_{match}(gram_n)$ is the maximum number of n-grams co-occurring in both a candidate summary and a set of reference summaries. ROUGE-N is a recall-based metric, as evidenced by the fact that the denominator in the equation represents the sum of n-grams present in the reference summary. Additionally, it should be noted that the numerator calculates the total sum of all reference summaries. This basically assigns greater importance to the occurrence of n-grams that appear in multiple references. Hence, the ROUGE-N measure favors a candidate summary that includes words that are common in many references. This is once again highly intuitive and logical, since we often favor a candidate summary that closely aligns with the majority among reference summaries [51].

Similar to conventional metrics, BERTSCORE calculates a similarity score for each token in the candidate sentence with respect to every token in the reference sentence [52]. As opposed to relying on exact matches, it calculates the similarity of tokens by utilizing contextual embeddings. It represents the tokens in contextual embeddings given a reference sentence and a candidate sentence, and calculate matching using cosine similarity, which electively be weighted with inverse document frequency scores. Fig. 9 depicts the procedure for computation.

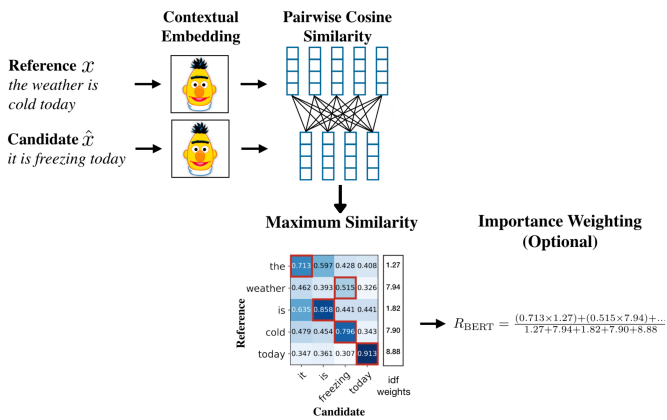

Fig. 9. Illustration of the computation of the recall metric R<sub>BERT</sub> [52].

Human judgement is the most prevailing and effective extrinsic evaluation technique. Linguists can assist in conducting evaluations using several criteria, including coherence, fluency, informativeness, relevance, and more. Nevertheless, conducting human evaluations would require a significant investment of time and financial resources, despite the high level of efficacy.

## VI. CONCLUSION

Through the perspective of extractive and abstractive categories, the author of this study examined a variety of text summarization classifications. In addition, the internal functioning of the approaches proposed in the existing literature were thoroughly investigated, ranging from simple statistical methods to complex transformer architectures. Finally, a comprehensive analysis of several evaluation methods for summarization has been conducted, ranging from basic metrics such as Euclidean distance to more advanced techniques like contextual embeddings based on BERTSCORE.

The task of text summarizing has made significant progress, but further enhancements are necessary, particularly in the field of abstractive text summarization. While transformers exhibit impressive performance, the challenge of addressing hallucinations [16] remains necessary. It would be very interesting to witness the advancements beyond the current accomplishments in this domain.

## REFERENCES

[1] E. Zolotareva, T. M. Tashu, and T. Horváth, "Abstractive Text Summarization using Transfer Learning".

[2] nasrin nazari and M. A. Mahdavi, "A survey on Automatic Text Summarization," *J. AI Data Min.*, no. Online First, May 2018, doi: 10.22044/jadm.2018.6139.1726.

[3] N. Moratanch and S. Chitrakala, "A survey on extractive text summarization," in *2017 International Conference on Computer, Communication and Signal Processing (ICCCSP)*, Chennai, India: IEEE, Jan. 2017, pp. 1–6. doi: 10.1109/ICCCSP.2017.7944061.

[4] M. Gambhir and V. Gupta, "Recent automatic text summarization techniques: a survey," *Artif. Intell. Rev.*, vol. 47, no. 1, pp. 1–66, Jan. 2017, doi: 10.1007/s10462-016-9475-9.

[5] P. Verma, A. Verma, and S. Pal, "An approach for extractive text summarization using fuzzy evolutionary and clustering algorithms," *Appl. Soft Comput.*, vol. 120, p. 108670, May 2022, doi: 10.1016/j.asoc.2022.108670.

[6] K. .D.K and D. M. N. Mubarak, "An Overview of Extractive Based Automatic Text Summarization Systems," *Int. J. Comput. Sci. Inf. Technol.*, vol. 8, no. 5, pp. 33–44, Oct. 2016, doi: 10.5121/ijcsit.2016.8503.

[7] W. S. El-Kassas, C. R. Salama, A. A. Rafea, and H. K. Mohamed, "Automatic text summarization: A comprehensive survey," *Expert Syst. Appl.*, vol. 165, p. 113679, Mar. 2021, doi: 10.1016/j.eswa.2020.113679.

[8] R. Mihalcea and P. Tarau, "TextRank: Bringing Order into Texts".

[9] G. Erkan and D. R. Radev, "LexRank: Graph-based Lexical Centrality as Salience in Text Summarization," *J. Artif. Intell. Res.*, vol. 22, pp. 457–479, Dec. 2004, doi: 10.1613/jair.1523.

[10] J. L. Neto, A. A. Freitas, and C. A. A. Kaestner, "Automatic Text Summarization Using a Machine Learning Approach," in *Advances in Artificial Intelligence*, vol. 2507, G. Bittencourt and G. L. Ramalho, Eds., in Lecture Notes in Computer Science, vol. 2507. , Berlin, Heidelberg: Springer Berlin Heidelberg, 2002, pp. 205–215. doi: 10.1007/3-540-36127-8_20.

[11] "'Extractive Text Summarization Using Machine Learning' by Swapnil Acharya." Accessed: Jun. 12, 2024. [Online]. Available: https://repository.stcloudstate.edu/csit_etds/39/

[12] N. S. Shirwandkar and S. Kulkarni, "Extractive Text Summarization Using Deep Learning," in *2018 Fourth International Conference on Computing Communication Control and Automation (ICCUBEA)*, Pune, India: IEEE, Aug. 2018, pp. 1–5. doi: 10.1109/ICCUBEA.2018.8697465.

[13] A. Rezaei, S. Dami, and P. Daneshjoo, "Multi-Document Extractive Text Summarization via Deep Learning Approach," in *2019 5th Conference on Knowledge Based Engineering and Innovation (KBEI)*, Tehran, Iran: IEEE, Feb. 2019, pp. 680–685. doi: 10.1109/KBEI.2019.8735084.

[14] B. Singh, R. Desai, H. Ashar, P. Tank, and N. Katre, "A Trade-off between ML and DL Techniques in Natural Language Processing," *J. Phys. Conf. Ser.*, vol. 1831, no. 1, p. 012025, Mar. 2021, doi: 10.1088/1742-6596/1831/1/012025.

[15] A. Vaswani *et al.*, "Attention Is All You Need." arXiv, Aug. 01, 2023. Accessed: Apr. 18, 2024. [Online]. Available: http://arxiv.org/abs/1706.03762

[16] J. Pilault, R. Li, S. Subramanian, and C. Pal, "On Extractive and Abstractive Neural Document Summarization with Transformer Language Models," in *Proceedings of the 2020 Conference on Empirical Methods in Natural Language Processing (EMNLP)*, Online: Association for Computational Linguistics, 2020, pp. 9308–9319. doi: 10.18653/v1/2020.emnlp-main.748.

[17] M. Zhang, G. Zhou, W. Yu, N. Huang, and W. Liu, "A Comprehensive Survey of Abstractive Text Summarization Based on Deep Learning," *Comput. Intell. Neurosci.*, vol. 2022, pp. 1–21, Aug. 2022, doi: 10.1155/2022/7132226.

[18] A. M. Rush, S. Chopra, and J. Weston, "A Neural Attention Model for Abstractive Sentence Summarization," 2015, doi: 10.48550/ARXIV.1509.00685.

[19] D. Suleiman and A. Awajan, "Deep Learning Based Abstractive Text Summarization: Approaches, Datasets, Evaluation Measures, and Challenges," *Math. Probl. Eng.*, vol. 2020, pp. 1–29, Aug. 2020, doi: 10.1155/2020/9365340.

[20] K. Lopyrev, "Generating News Headlines with Recurrent Neural Networks," 2015, doi: 10.48550/ARXIV.1512.01712.

[21] J. Cheng, L. Dong, and M. Lapata, "Long Short-Term Memory-Networks for Machine Reading." arXiv, Sep. 20, 2016. Accessed: Apr. 22, 2024. [Online]. Available: http://arxiv.org/abs/1601.06733

[22] Preethi. S, K. Shibi. M.S, Sheshan. S, R. Kingsy Grace, and M. Sri Geetha, "Abstractive Summarizer using Bi-LSTM," in *2022 International Conference on Edge Computing and Applications (ICECAA)*, Tamilnadu, India: IEEE, Oct. 2022, pp. 1605–1609. doi: 10.1109/ICECAA55415.2022.9936215.

[23] S. Song, H. Huang, and T. Ruan, "Abstractive text summarization using LSTM-CNN based deep learning," *Multimed. Tools Appl.*, vol. 78, no. 1, pp. 857–875, Jan. 2019, doi: 10.1007/s11042-018-5749-3.

[24] N. Dang, A. Khanna, and V. R. Allugunti, "TS-GAN with Policy Gradient for Text Summarization," in *Data Analytics and Management*, vol. 54, A. Khanna, D. Gupta, Z. Pólkowski, S. Bhattacharyya, and O. Castillo, Eds., in Lecture Notes on Data Engineering and

[24] Communications Technologies, vol. 54. , Singapore: Springer Singapore, 2021, pp. 843–851. doi: 10.1007/978-981-15-8335-3_64.

[25] K. Yao, L. Zhang, D. Du, T. Luo, L. Tao, and Y. Wu, "Dual Encoding for Abstractive Text Summarization," *IEEE Trans. Cybern.*, vol. 50, no. 3, pp. 985–996, Mar. 2020, doi: 10.1109/TCYB.2018.2876317.

[26] L. Hou, P. Hu, and C. Bei, "Abstractive Document Summarization via Neural Model with Joint Attention," in *Natural Language Processing and Chinese Computing*, vol. 10619, X. Huang, J. Jiang, D. Zhao, Y. Feng, and Y. Hong, Eds., in Lecture Notes in Computer Science, vol. 10619. , Cham: Springer International Publishing, 2018, pp. 329–338. doi: 10.1007/978-3-319-73618-1_28.

[27] Md. M. Rahman and F. H. Siddiqui, "Multi-layered attentional peephole convolutional LSTM for abstractive text summarization," *ETRI J.*, vol. 43, no. 2, pp. 288–298, Apr. 2021, doi: 10.4218/etrij.2019-0016.

[28] T. Vo, "SGAN4AbSum: A Semantic-Enhanced Generative Adversarial Network for Abstractive Text Summarization." Jul. 30, 2021. doi: 10.21203/rs.3.rs-648146/v1.

[29] S. Gao *et al.*, "Abstractive Text Summarization by Incorporating Reader Comments," *Proc. AAAI Conf. Artif. Intell.*, vol. 33, no. 01, pp. 6399–6406, Jul. 2019, doi: 10.1609/aaai.v33i01.33016399.

[30] V. M. Ratianantitra, J. L. Razafindramintsa, T. Mahatody, C. Rasoamalalavao, and V. Manantsoa, "Malagasy Abstractive Text Summarization Using Scheduled Sampling Model:," presented at the 2nd International Conference on Industry 4.0 and Artificial Intelligence (ICIAI 2021), Sousse, Tunisia, 2022. doi: 10.2991/aisr.k.220201.002.

[31] K. Ganesan, C. Zhai, and J. Han, "Opinosis: A Graph Based Approach to Abstractive Summarization of Highly Redundant Opinions," 2010, Accessed: Jun. 14, 2024. [Online]. Available: https://hdl.handle.net/2142/16555

[32] Z. Liang, J. Du, Y. Shao, and H. Ji, "Gated Graph Neural Attention Networks for abstractive summarization," *Neurocomputing*, vol. 431, pp. 128–136, Mar. 2021, doi: 10.1016/j.neucom.2020.09.066.

[33] Z. Song and I. King, "Hierarchical Heterogeneous Graph Attention Network for Syntax-Aware Summarization," *Proc. AAAI Conf. Artif. Intell.*, vol. 36, no. 10, pp. 11340–11348, Jun. 2022, doi: 10.1609/aaai.v36i10.21385.

[34] P. Kouris, G. Alexandridis, and A. Stafylopatis, "Text summarization based on semantic graphs: An abstract meaning representation graph-to-text deep learning approach." Aug. 11, 2022. doi: 10.21203/rs.3.rs-1938526/v1.

[35] E. Baralis, L. Cagliero, N. Mahoto, and A. Fiori, "GraphSum: Discovering correlations among multiple terms for graph-based summarization," *Inf. Sci.*, vol. 249, pp. 96–109, Nov. 2013, doi: 10.1016/j.ins.2013.06.046.

[36] S. Arumugam and S. B. Subramani, "Similitude Based Segment Graph Construction and Segment Ranking for Automatic Summarization of Text Document," *Trends Sci.*, vol. 19, no. 1, p. 1719, Jan. 2022, doi: 10.48048/tis.2022.1719.

[37] H. Yadav, N. Patel, and D. Jani, "Fine-Tuning BART for Abstractive Reviews Summarization," in *Computational Intelligence*, vol. 968, A. Shukla, B. K. Murthy, N. Hasteer, and J.-P. Van Belle, Eds., in Lecture Notes in Electrical Engineering, vol. 968. , Singapore: Springer Nature Singapore, 2023, pp. 375–385. doi: 10.1007/978-981-19-7346-8_32.

[38] J. Zhang, Y. Zhao, M. Saleh, and P. Liu, "PEGASUS: Pre-training with Extracted Gap-sentences for Abstractive Summarization," in *Proceedings of the 37th International Conference on Machine Learning*, PMLR, Nov. 2020, pp. 11328–11339. Accessed: Apr. 22, 2024. [Online]. Available: https://proceedings.mlr.press/v119/zhang20ae.html

[39] C. Raffel *et al.*, "Exploring the Limits of Transfer Learning with a Unified Text-to-Text Transformer." arXiv, Sep. 19, 2023. Accessed: Apr. 22, 2024. [Online]. Available: http://arxiv.org/abs/1910.10683

[40] M. Guo *et al.*, "LongT5: Efficient Text-To-Text Transformer for Long Sequences." arXiv, May 03, 2022. Accessed: Apr. 18, 2024. [Online]. Available: http://arxiv.org/abs/2112.07916

[41] S. Kim and B. Yoon, "Multi-document summarization for patent documents based on generative adversarial network," *Expert Syst. Appl.*, vol. 207, p. 117983, Nov. 2022, doi: 10.1016/j.eswa.2022.117983.

[42] P. J. Liu *et al.*, "Generating Wikipedia by Summarizing Long Sequences." arXiv, Jan. 30, 2018. doi: 10.48550/arXiv.1801.10198.

[43] U. Khandelwal, K. Clark, D. Jurafsky, and L. Kaiser, "Sample Efficient Text Summarization Using a Single Pre-Trained Transformer." arXiv, 2019. doi: 10.48550/ARXIV.1905.08836.

[44] L. Phan, H. Tran, H. Nguyen, and T. H. Trinh, "ViT5: Pretrained Text-to-Text Transformer for Vietnamese Language Generation." arXiv, 2022. doi: 10.48550/ARXIV.2205.06457.

[45] P. Grouchy *et al.*, "An Experimental Evaluation of Transformer-based Language Models in the Biomedical Domain." arXiv, 2020. doi: 10.48550/ARXIV.2012.15419.

[46] Computer Science Department, School of Computer Science, Bina Nusantara University, Jakarta Indonesia 11480, J. L. Atipa, J. Javin, F. Bryan, V. Yesmaya, and R. Wongso, "Abstractive Text Summary with Transformer on Youtube Video Subtitle," *Int. J. Emerg. Technol. Adv. Eng.*, vol. 13, no. 2, pp. 1–7, Feb. 2023, doi: 10.46338/ijetae0223_01.

[47] S. Ghosh Roy, N. Pinnaparaju, R. Jain, M. Gupta, and V. Varma, "Summaformers @ LaySumm 20, LongSumm 20," in *Proceedings of the First Workshop on Scholarly Document Processing*, Online: Association for Computational Linguistics, 2020, pp. 336–343. doi: 10.18653/v1/2020.sdp-1.39.

[48] M. Jiang, Y. Zou, J. Xu, and M. Zhang, "GATSum: Graph-Based Topic-Aware Abstract Text Summarization," *Inf. Technol. Control*, vol. 51, no. 2, pp. 345–355, Jun. 2022, doi: 10.5755/j01.itc.51.2.30796.

[49] J. Steinberger and K. Ježek, "Evaluation Measures for Text Summarization," *Comput. Inform.*, vol. 28, no. 2, Art. no. 2, 2009, Accessed: Jun. 15, 2024. [Online]. Available: https://www.cai.sk/ojs/index.php/cai/article/view/37

[50] P. Xia, L. Zhang, and F. Li, "Learning similarity with cosine similarity ensemble," *Inf. Sci.*, vol. 307, pp. 39–52, Jun. 2015, doi: 10.1016/j.ins.2015.02.024.

[51] C.-Y. Lin, "ROUGE: A Package for Automatic Evaluation of Summaries".

[52] T. Zhang, V. Kishore, F. Wu, K. Q. Weinberger, and Y. Artzi, "BERTScore: Evaluating Text Generation with BERT," 2019, doi: 10.48550/ARXIV.1904.09675.